%% file: neurips_2025.tex
\documentclass{article}

\usepackage[main, preprint]{neurips_2025}

\usepackage[utf8]{inputenc} 
\usepackage[T1]{fontenc}    
\usepackage{hyperref}
\usepackage{url}            
\usepackage{booktabs}       
\usepackage{amsmath}        
\usepackage{amssymb}        
\usepackage{amsfonts}       
\usepackage{nicefrac}       
\usepackage{microtype}      
\usepackage{xcolor}         
\usepackage{natbib}         
\usepackage{graphicx}
\usepackage{float}
\usepackage{algorithm}
\usepackage{algpseudocode}
\usepackage{subcaption}

\usepackage{xcolor}

\usepackage[most]{tcolorbox}
\newtcolorbox{takeawaybox}{
  colback=red!5!white,       
  colframe=black,            
  boxrule=0.6pt,
  arc=2mm,
  left=6pt,right=6pt,top=4pt,bottom=4pt
}

\usepackage{booktabs}
\usepackage{multirow}
\usepackage{makecell}
\usepackage{xcolor}
\usepackage{graphicx}
\usepackage{siunitx}
\usepackage{pifont}

\usepackage{wrapfig}
\usepackage{booktabs}
\usepackage{multirow}

\newcommand{\cmark}{\ding{51}}

\title{The Cancellation Hypothesis in Critic-Free RL: \\ From Outcome Rewards to Token Credits}

\author{%
  \textbf{Tianhao Cheng\textsuperscript{1,2}\thanks{Equal contribution.} \quad Zeyu Huang\textsuperscript{3}\footnotemark[1] \quad Zihan Qiu\textsuperscript{6}} \\
  \textbf{Yu Cheng\textsuperscript{2,4} \quad Edoardo Ponti\textsuperscript{3} \quad Yinghui Xu\textsuperscript{1} \quad Ivan Titov\textsuperscript{3,7} \quad Zenglin Xu\textsuperscript{1,5}} \\
  \textsuperscript{1}Fudan University \quad \textsuperscript{2}Shanghai Innovation Institute \quad \textsuperscript{3}The University of Edinburgh \\
  \textsuperscript{4}The Chinese University of Hong Kong \quad \textsuperscript{5}Shanghai Academy of AI for Science \\
  \textsuperscript{6}Qwen Team, Alibaba Group \quad \textsuperscript{7}University of Amsterdam \\
}

\begin{document}

\maketitle

\begin{abstract}
A commonly accepted explanation of critic-free RL (e.g., GRPO) for LLMs, based on sequence-level rewards, is that it reinforces successful rollouts with a positive advantage while penalizing failed ones. In contrast, we study critic-free RL from a token-level perspective, revealing the \emph{token-flipping} phenomenon: positive and negative rollouts exhibit remarkably similar proportions of tokens whose probabilities are boosted or suppressed during RL training.
To explain this phenomenon, we further show that a token's change in probability is not fully determined by its own advantage; coupled gradient interactions with other tokens also play a non-negligible role.
Specifically, these \textit{token coupling effects} occur primarily between identical tokens that are both predicted with low confidence.
Building upon this analysis, we propose the \emph{cancellation hypothesis}: as a result of coupling, opposing signals cancel out for tokens shared by positive and negative rollouts, while tokens more specific to successful rollouts receive stronger reinforcement, thereby inducing hidden token-level credit assignment from rollout-level rewards.
We support this hypothesis with complementary empirical evidence.
(1) Compared with training on only positive rollouts, critic-free RL shifts updates from template and formatting tokens toward reasoning tokens;
(2) Tokens boosted by critic-free RL consistently demonstrate higher value than suppressed tokens, regardless of whether they originate from positive or negative rollouts.
Guided by this view, we implement two batching interventions to encourage or preserve cancellation in critic-free RL training: query-preserved mini-batching, which keeps rollouts from the same prompt together to encourage token coupling, and reward-balanced batching, which ensures that positive and negative rollouts are balanced in each update.
Despite their simplicity, these interventions improve RLVR training across multiple model scales, supporting cancellation as both an explanatory principle and a practical design criterion for critic-free RL training.
\end{abstract}

\section{Introduction}
Reinforcement learning with verifiable rewards (RLVR) has become one of the standard recipes for post-training large language models (LLMs)~\citep{guo2025deepseek,jaech2024openai,team2025kimi}.
Among RLVR methods, critic-free approaches such as Group Relative Policy Optimization (GRPO)~\citep{shao2024deepseekmath} and its variants~\citep{yu2025dapo,liu2025understanding} are widely adopted for their simplicity.
They remove the value network~\citep{schulman2017proximal} and instead assign a single, shared advantage to every token within a rollout.
Despite such coarse, rollout-level supervision, these methods achieve strong results across reasoning and broader agentic domains~\citep{guo2025deepseek}.
This success motivated growing interest in understanding how critic-free RL optimizes and reshapes model behavior.
A common view is that RL reinforces successful rollouts and suppresses failed ones, thereby shifting the model distribution toward higher-reward responses~\citep{mroueh2025reinforcement, wu2025takes}.
Recent work complicates this view by analyzing the different contributions of positive and negative rollouts to optimization dynamics~\citep{zhu2025surprising,tang2025rethinking,deng2025effect,deng2025token,deng2025grpo,simoni2025gtpo,deng2025grpo,xue2025simpletir}.

However, a fundamental disconnect remains in the current discourse: most existing analyses focus on rollout-level likelihoods or outcomes, while treating the token-level dynamics as a black box.
Though the model improves, it still lacks a mechanistic to understand how a shared rollout-level advantage distributes credit or blame across tokens.
Under the \emph{de-facto} sequence-level intuition, tokens in successful rollouts should be mostly reinforced, and vice versa.

\begin{figure}[t]
\vspace{-20pt}
  \centering
  \includegraphics[width=\linewidth]{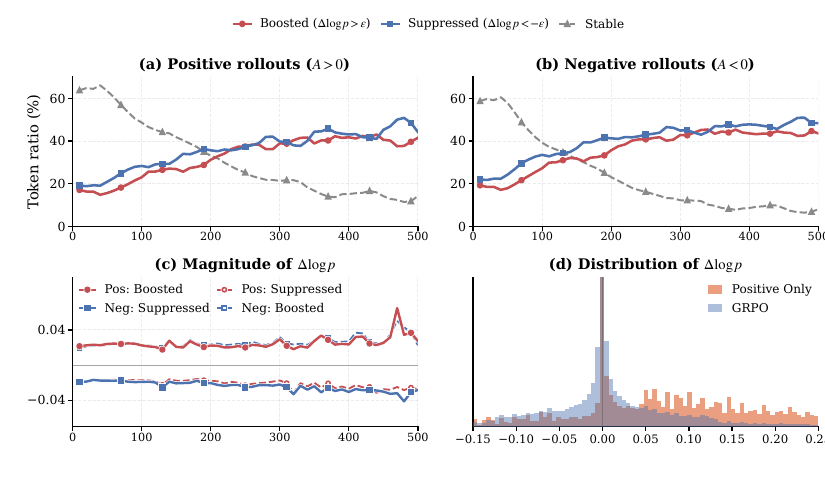}
  \vspace{-20pt}
  \caption{Token-level updates do not simply follow rollout-level advantage signs. Tokens could be categorized into three groups according to their change in log probability: boosted, suppressed, and stable. Figures (a) and (b) show that positive and negative rollouts exhibit remarkably similar boosted/suppressed/stable token ratios during training. Figure (c) shows that the update magnitudes are also similar. Figure (d) further confirms that the flipping pattern greatly decreases when training with only positive rollouts, highlighting the coupling effect between positive and negative rollouts.
  }
  \label{fig:push_ratio_posneg}
    \vspace{-15pt}
\end{figure}

In this work, we bridge this gap by shifting the analytical lens from rollouts to tokens.
We measure the shift in log-probability for every token before and after individual GRPO training step, as shown in Fig.~\ref{fig:push_ratio_posneg}(a,b).
Our findings reveal a clear departure from conventional intuition: tokens in successful rollouts are not mostly reinforced, nor are tokens in failed rollouts mostly suppressed.
Instead, nearly half of the tokens move in the direction opposite to what their rollout-level advantage sign suggests.
More surprisingly, positive and negative rollouts exhibit remarkably similar ratios of boosted and suppressed tokens during training, even the magnitudes of these token-level changes are also similar (Fig.~\ref{fig:push_ratio_posneg}(c)).
We term this mismatch \emph{Token Flipping}, which suggests that the rollout-level advantage sign is a poor predictor for token-level updates, implying that the learning mechanism in critic-free RL is more granular than the standard sequence-level view suggests.
Furthermore, this pattern largely disappears under positive-rollout-only training, where negative rollouts are masked (Fig.~\ref{fig:push_ratio_posneg}(d)), indicating that token flipping arises from interactions between positive and negative rollouts.

The \emph{Token Flipping} phenomenon suggests that a token's update cannot be explained by its advantage alone.
To understand where the additional force arises, we revisit GRPO's training objective at the token level.
A first-order expansion shows that the log-probability change of token $j$ can be decomposed into a self-term and a within-batch coupling term.
The former depends on its own advantage, while the latter aggregates interactions with all other tokens within the batch.
We then ask when such coupling is large enough to affect the update.
By approximating the full gradient with an output-layer proxy, we obtain an interpretable factorization of the coupling kernel into a representation term and a distributional term.
This factorization reveals a sparse structure: non-negligible coupling occurs primarily between identical tokens that are both predicted with low confidence.
This helps explain why token-level updates could deviate from their own advantage signs, and suggests that the joint composition of an RL batch plays a central role in shaping the effective learning signal.

\begin{figure}[t]
  \vspace{-30pt}
  \centering
    \includegraphics[width=\linewidth]{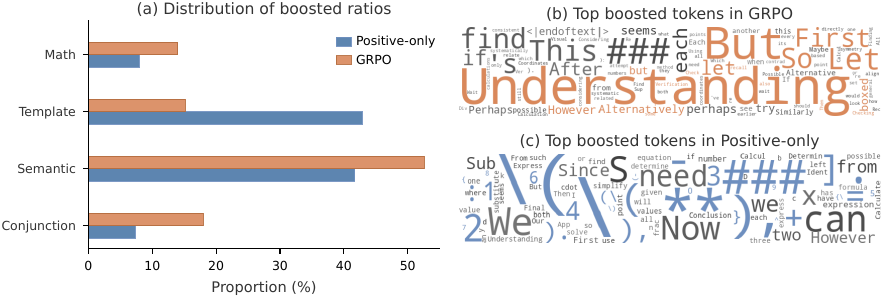}
    \caption{
      (a) Tokens are grouped by NLTK POS tagging into Math (Number, Latex), Template (Symbol, Determiner, Conj), Semantic (Noun, Verb), and Conjunction (Conjunction); Each bar shows the fraction of total log-prob boost contributed by that group. Positive-only updates concentrate on Template, whereas GRPO update toward Semantic and Conjunction.                                                      
      (b/c) At the token level, Positive-only boosts formatting tokens, GRPO boosts reasoning conjunctions.
    }
  \label{fig:cancellation_denoising}
\vspace{-15pt}
\end{figure}

Building on this sparse coupling structure, we propose the \emph{cancellation hypothesis}: tokens that appear in both positive and negative rollouts receive opposing coupled signals that cancel in the net update, while tokens more specific to successful rollouts retain stronger reinforcement.
In this way, batch-level token coupling can induce hidden token-level credit assignment from rollout-level rewards, even without an explicit critic:
(1) compared with training on only positive rollouts, GRPO greatly weakens updates on template and formatting tokens and shifts emphasis toward reasoning tokens (Fig.~\ref{fig:cancellation_denoising}).
(2) we estimate token value by Monte Carlo rollouts and find that GRPO-boosted tokens consistently have higher estimated value than suppressed tokens, regardless of whether they originate from positive or negative rollouts (Fig.~\ref{fig:value_combined_row}(a)).
This value gap further peaks in an intermediate-entropy region, connecting cancellation to entropy-based token selection~\citep{wang2025beyond}.

Finally, the cancellation view leads us to implement two simple yet effective interventions for RL training.
(1) \emph{query-preserved mini-batching}, which keeps rollouts from the same query together to ensure related positive and negative samples are within the same optimizer step.
(2) \emph{reward-balanced batching}, which avoids batches that are overwhelmed by positive or negative rollouts to preserve the cancellation structure.
Across multiple model scales and math benchmarks, these interventions improve RLVR training, demonstrating the practical value of preserving the cancellation hypothesis.

\section{Token Coupling Effect}

The \emph{Token Flipping} suggests that a token's update direction may be shaped by other tokens in the same batch, rather than by its own advantage alone.
In this section, we formalize this batch-level token coupling for a single RL update.
Unless stated otherwise, we consider GRPO with binary rewards: for each prompt $q$, a group of $G$ responses $\{o_i\}_{i=1}^G$ is sampled from the old policy $\pi_{\theta_{\mathrm{old}}}$, and each response is assigned a rollout-level advantage $\hat A_i$ after verification.
We study the change in the token's log-probability before and after an update step.
For the $t$-th token in response $i$, denoted $o_{i,t}$,
\begin{equation}
\Delta\log p_{i,t}
\triangleq
\log \pi_{\theta'}\!\left(o_{i,t}\mid q, o_{i,<t}\right)
-
\log \pi_{\theta_{\mathrm{old}}}\!\left(o_{i,t}\mid q, o_{i,<t}\right).
\label{eq:token_displacement1}
\end{equation}
For this analysis, we rewrite the GRPO update as a weighted token log-likelihood:
\begin{equation}
  \mathcal{J}(\theta)
  \;\triangleq\;
  \frac{1}{N}\sum_{i=1}^{G}\sum_{t=1}^{|o_i|}
  A_i \log \pi_{\theta}(o_{i,t}\mid q, o_{i,<t}),
  \qquad
  N \triangleq \sum_{i=1}^{G}|o_i|.
  \label{eq:simplified_grpo_obj}
\end{equation}
$A_i$ denotes the update weight obtained from the original rollout advantage $\hat A_i$. As the importance-sampling ratio and clipping term only rescale or mask the gradient, we absorb them into $A_i$.
Under a first-order approximation, a single gradient step on $\mathcal{J}(\theta)$ with step size $\eta$ yields
\begin{equation}
\theta' \;=\; \theta + \eta \nabla_\theta \mathcal{J}(\theta)
\;=\;
\theta + \frac{\eta}{N}\sum_{i=1}^{G}\sum_{t=1}^{|o_i|}
A_i \, \nabla_\theta \log \pi_{\theta}(o_{i,t}\mid q, o_{i,<t}).
\label{eq:theta_update}
\end{equation}

We now analyze the $\Delta\log p_{i,t}$ under the one-step update:
let $g_{i,t} \triangleq \nabla_\theta \log \pi_{\theta}(o_{i,t}\mid q, o_{i,<t})$ denote the score function with respect to full parameters $\theta$, a first-order Taylor expansion gives
\begin{equation}
\Delta\log p_{i,t}
\;\approx\;
\nabla_\theta \log \pi_\theta(o_{i,t}\mid q,o_{i,<t})^\top (\theta' - \theta)
\;=\;
g_{i,t}^{\top}(\theta'-\theta).
\label{eq:taylor}
\end{equation}

Substituting Eq.~\eqref{eq:theta_update} into Eq.~\eqref{eq:taylor} and flattening tokens in the batch into indices $j,k \in \{1,\dots,N\}$:
\begin{align}
\Delta \log p_j \approx \frac{\eta}{N}\sum_{k=1}^{N} A_k \langle g_j, g_k \rangle
= \frac{\eta}{N}\left( A_j \lVert g_j \rVert_2^2 + \sum_{k \ne j} A_k \langle g_j, g_k \rangle \right)
\label{eq:coupling_main}
\end{align}
where $A_k$ denotes the effective weight inherited from the rollout of token $k$.
Eq.~\ref{eq:coupling_main} clearly reveals that the update of token $j$ is not only driven by
the advantage $A_j$, but also the coupling term $\sum_{k \ne j} A_k \langle g_j, g_k \rangle$.
We define the coupling kernel $K_{j,k} \triangleq \langle g_j, g_k\rangle$, which determines whether the couping effect between the token $i$ and token $j$ could be non-negligible? or not.
Computing this kernel in the full parameter space is expensive, since it requires inner products between full-model gradients.
We therefore follow~\citet{deng2025effect} and adopt the \emph{output-layer proxy}, which approximates $g_j$ using only the gradient with respect to the final unembedding matrix, denoted $g_j^{(W)}$.
Appendix~\ref{appendix:proxy} shows that this proxy closely tracks the full-parameter update in practice.
Let $z_\theta(s)=Wh_\theta(s)$ denote the logits, where $h_\theta(s)$ is the final hidden state, the gradient with respect to $W$ becomes
\begin{equation}
g_j^{(W)} 
\;\triangleq\; \nabla_W \log \pi_\theta(o_j \mid s_j) 
\;=\; \underbrace{(\nabla_z \log \pi_\theta(o_j \mid s_j))}_{\text{error term } r_j} \underbrace{h_j^\top}_{\text{input}}
\;=\; r_j h_j^\top,
\label{eq:score_outer_product}
\end{equation}
where $s_j$ is the prefix state of token $j$, $h_j=h_\theta(s_j)$, and $r_j \triangleq e_{o_j} - \pi_\theta(\cdot \mid s_j) \in \mathbb{R}^{|\mathcal{V}|}$.
Substituting Eq.~\eqref{eq:score_outer_product} into the kernel definition and using the Frobenius inner product property $\langle u v^\top, x y^\top \rangle_F = \langle u, x \rangle \langle v, y \rangle$, we obtain the output-layer proxy factorization:
\begin{equation}
K_{j,k} 
\;\approx\; \langle g_j^{(W)}, g_k^{(W)} \rangle_F 
\;=\; \langle r_j h_j^\top, r_k h_k^\top \rangle_F 
\;=\; \underbrace{\langle h_j, h_k \rangle}_{\text{Representation}} \cdot \underbrace{\langle r_j, r_k \rangle}_{\text{Distribution}}.
\label{eq:kernel_decoupling}
\end{equation}
Thus, non-negligible coupling requires both aligned hidden representations and high similarity between the output distributions.
We denote the second factor by $\phi_{j,k} \triangleq \langle r_j, r_k \rangle$ and expand it:
\begin{equation}  
  \phi_{j,k} 
  \;=\; (e_{o_j} - \pi_j)^\top (e_{o_k} - \pi_k) 
  \;=\; \mathbb{I}[o_j=o_k] - \pi_j(o_k) - \pi_k(o_j) + \langle \pi_j, \pi_k \rangle.
  \label{eq:phi_expansion}
\end{equation}

Substituting this factorized kernel back into Eq.~\ref{eq:coupling_main}, we obtain the proxy delta:
\begin{equation}
\delta_j \approx \frac{\eta}{N}\sum_{k=1}^{N} A_k \langle h_j, h_k\rangle \phi_{j,k}
=
\frac{\eta}{N}\left(
A_j \langle h_j, h_j\rangle \phi_{j,j}
+
\sum_{k\neq j} A_k \langle h_j, h_k\rangle \phi_{j,k}
\right).
\label{eq:self_cross_force}
\end{equation}
We now inspect $\phi_{j,k}$ in two cases:
\begin{equation}
\phi_{j,k}
=
\begin{cases}
- \pi_j(o_k) - \pi_k(o_j) + \langle \pi_j, \pi_k \rangle, & o_j \neq o_k, \\[4pt]
(1 - \pi_j(o))(1 - \pi_k(o)), & o_j = o_k = o.
\end{cases}
\label{eq:uncertainty_factorization}
\end{equation}
For distinct output tokens $o_j \neq o_k$, the cross-token probabilities $\pi_j(o_k)$ and $\pi_k(o_j)$ are usually small, and the overlap term $\langle \pi_j,\pi_k\rangle$ is also small when the two output distributions have weak overlap.
Thus, different-token pairs typically contribute little through $\phi_{j,k}$.
For same-token pairs $o_j=o_k=o$, however, $\phi_{j,k}=(1-\pi_j(o))(1-\pi_k(o))$ is positive and becomes large when both tokens are predicted with low confidence.
Therefore, the similarity between output distributions makes coupling sparse: non-negligible coupling occurs primarily between identical, low-confidence tokens.

\begin{figure}[t]
\vspace{-20pt}
  \centering
    \includegraphics[width=\linewidth]{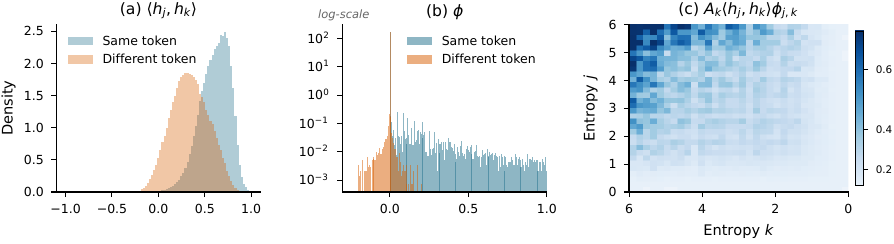}
    \caption{Coupling concentrates on same-token, low-confidence pairs. \textbf{(a)} Hidden-representation similarities $\langle h_j,h_k\rangle$ are generally positive across token pairs, with same-token pairs showing stronger alignment than different-token pairs. \textbf{(b)} The output-distribution factor $\phi_{j,k}$ is non-negligible primarily for same-token pairs. \textbf{(c)} The full coupling contribution $A_k\langle h_j,h_k\rangle\phi_{j,k}$ is strongest when both tokens have high entropy, corresponding to low-confidence predictions.}
    \vspace{-20pt}
  \label{fig:coupling_sparsity}
\end{figure}

\paragraph{Empirical Verification}
We first look into the similarity of hidden representation and output distribution between sampled tokens in Eq.~\ref{eq:kernel_decoupling} and the sparsity pattern predicted by Eq.~\ref{eq:uncertainty_factorization}.
As shown in Fig.~\ref{fig:coupling_sparsity}(a), hidden-representation similarities are generally positive across token pairs, and same tokens tend to be aligned more strongly than different-token pairs.
Fig.~\ref{fig:coupling_sparsity}(b) further shows that non-negligible $\phi_{j,k}$ values are primarily on the same-token pairs, though most of them are zero.
Finally, Fig.~\ref{fig:coupling_sparsity}(c) shows that the full coupling contribution is strongest when both tokens have high entropy, corresponding to low-confidence predictions.
Together, these observations support our factorization: \emph{non-negligible coupling is sparse and occurs primarily between identical, low-confidence tokens.}

\begin{figure}[t]
\vspace{-20pt}
  \centering
    \includegraphics[width=\linewidth]{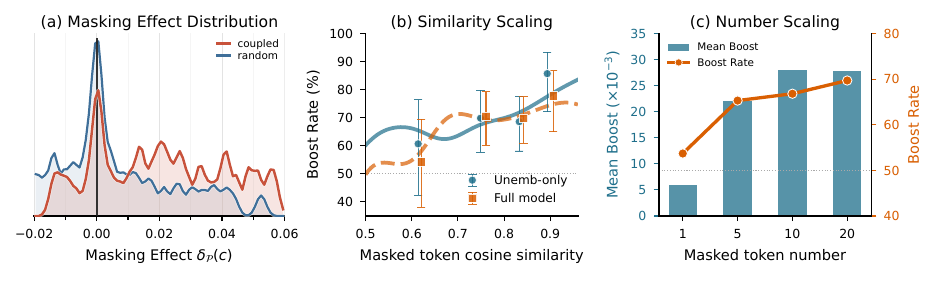}
    \caption{
    Empirical verification via masked updates.
    \textbf{(a)} Coupled tokens produce a right-shifted Masking Effect distribution compared with random tokens.
    \textbf{(b)} Boost Rate increases with coupling strength and is consistent between the output-layer proxy and the full-parameter update.
    \textbf{(c)} Masking more coupled tokens increases both Mean Boost and Boost Rate.
    }
  \label{fig:causal_study}
\vspace{-10pt}
\end{figure}

For further verification, we test whether tokens selected by this coupling structure measurably affect actual updates.
Specifically, we compare the normal update with a masked update, where a selected token set $\mathcal{M}(c)$ is removed from the loss.
We define the \emph{Masking Effect} as
\[
\delta_{\mathcal{M}}(c)
=
\log \pi_{\theta(c\mid s_c)}
-
\log \pi_{\theta}^{-\mathcal{M}}(c\mid s_c),
\]
where $\log \pi_{\theta}^{-\mathcal{M}}(c\mid s_c)$ is the log probability with the masked update.
A positive value of $\delta$ indicates that the removed tokens would have increased the candidate token's log-probability under the unmasked update.
We summarize this effect with two statistics: \emph{Boost Rate}, the fraction of candidates for which the selected tokens boost the candidate token's log-probability, and \emph{Mean Boost}, the average Masking Effect:
\[
  \mathrm{Boost\ Rate}=\frac{1}{|\mathcal{C}|}\sum_{c\in\mathcal{C}}\mathbf{1}\{\delta_{\mathcal{M}(c)}(c)>0\},\qquad
  \mathrm{Mean\ Boost}=\frac{1}{|\mathcal{C}|}\sum_{c\in\mathcal{C}}\delta_{\mathcal{M}(c)}(c).
\]
\begin{wraptable}{r}{0.45\linewidth}
  \vspace{-1.2em}
  \centering
  \scriptsize
  \setlength{\tabcolsep}{2.0pt}
  \renewcommand{\arraystretch}{0.88}
  \setlength{\abovecaptionskip}{2pt}
  \setlength{\belowcaptionskip}{2pt}
  \caption{
    Masked updates verify the effect of coupled tokens.
    Tokens satisfying both same-token and low-confidence conditions yield higher Boost Rate and larger Mean Boost than random tokens under both the output-layer proxy and full-parameter update.
  }
  \label{tab:directional_masking}
  \vspace{-0.5em}
  \begin{tabular*}{\linewidth}{@{}c@{\hspace{3pt}}c@{\extracolsep{\fill}}r@{\hspace{2pt}}r@{\extracolsep{\fill}}r@{\hspace{2pt}}r@{}}
  \toprule
  \multicolumn{2}{c}{Masked Tokens}
  & \multicolumn{2}{c}{LM-head}
  & \multicolumn{2}{c}{Full} \\
  \cmidrule(lr){1-2} \cmidrule(lr){3-4} \cmidrule(lr){5-6}
  Same & LowConf. & \makecell{Boost\\Rate} & \makecell{Mean\\Boost} & \makecell{Boost\\Rate} & \makecell{Mean\\Boost} \\
  \midrule
       &        & 50.60 & $-.0039$ & 50.21 & $+.0048$ \\
  \cmark &      & 66.20 & $+.0224$ & 66.41 & $+.0241$ \\
     & \cmark & 44.20 & $-.0004$ & 43.04  & $-.0012$ \\
  \cmark & \cmark & 74.12 & $+.0301$ & 75.00 & $+.0306$ \\
  \bottomrule
  \end{tabular*}
  \vspace{-1.0em}
\end{wraptable}
As shown in Table~\ref{tab:directional_masking}, masking random tokens yields a random Boost Rate ($\approx$50\%) and a near-zero Mean Boost.
In contrast, masking tokens that satisfy both conditions, same token identity and low confidence, raises Boost Rate to approximately 75\% and produces clearly positive Mean Boost under both the output-layer proxy and the full-parameter update.
This verifies that the sparse coupling structure identified by Eq.~\ref{eq:uncertainty_factorization} selects tokens with a consistent and measurable influence on the candidate update.
The close agreement between output-layer and full-parameter update further suggests that the proxy approximates the full-parameter training dynamics well.
Fig.~\ref{fig:causal_study} provides a complementary view, showing that Masking Effect scales with predicted coupling strength and is not driven by a few outlier pairs.

\section{Coupling Effect Induces Cancellation}
The previous section indicates that token updates are not isolated: same-token, low-confidence, and representation-aligned tokens can exert coupled influence on one another within the same batch.
In a group-relative batch, such coupled directions often appear in both successful and failed rollouts from the same query.
Because these rollouts carry opposite-signed group-normalized advantages, shared directions are suppressed in the aggregate update, while directions that are more correlated with successful rollouts are reinforced.
We term this mechanism the \emph{cancellation hypothesis}.
It has two consequences: it could reduce variance along shared directions, and it turns the group-relative update into an advantage-weighted filter that induces hidden token-level credit assignment.

\subsection{Cancellation as Advantage-Weighted Filter and Variance Reducer}
\label{sec:structured_cancellation}
Consider a query $q$ with $G$ sampled rollouts and group-normalized rollout advantages $\{\hat A_i\}_{i=1}^G$ satisfying $\sum_{i=1}^G \hat A_i=0$.
Let $d_i=\sum_t g_{i,t}$ denote the response-level score direction for rollout $i$, where $g_{i,t}=\nabla_\theta \log \pi_\theta(o_{i,t}\mid q,o_{i,<t})$.
The group-relative part of the query-level update is $\Delta \theta_q \propto \sum_{i=1}^G \hat A_i d_i$. The squared norm of the query-level update could be expanded:
\begin{equation}
\left\|\sum_{i=1}^G \hat A_i g_i\right\|^2
=
\sum_{i=1}^G \hat A_i^2\|g_i\|^2
+
\sum_{i\ne j}\hat A_i\hat A_j\langle g_i,g_j\rangle .
\label{eq:group_norm_expansion}
\end{equation}
As rollouts from the same query contain  identical tokens and similar hidden states, their response-level gradients have more positive overlap.
In the idealized case where the common within-query overlap is approximately a positive constant $c_q$, the cross term becomes
\begin{equation}
\sum_{i\ne j}\hat A_i\hat A_j\langle g_i,g_j\rangle
\approx
c_q\sum_{i\ne j}\hat A_i\hat A_j
=
-c_q\sum_{i=1}^G \hat A_i^2 < 0,
\label{eq:negative_cross_term}
\end{equation}
where the last equality uses $\sum_i \hat A_i=0$.
Thus, positive gradient overlap is multiplied by negative advantage cross-products, reducing the update norm relative to the self-term $\sum_i \hat A_i^2\|g_i\|^2$.
Since gradient variance is controlled by the second moment of the update, this norm reduction gives the variance-reduction role of cancellation.

The same algebra also gives a filtering interpretation.
For any token or feature direction $u$, let $m_{i,u}=\langle g_i,u\rangle$ measure how strongly rollout $i$ loads on that direction.
The surviving signal along $u$ is
\begin{equation}
S_u = \sum_{i=1}^G \hat A_i m_{i,u}
=
\sum_{i=1}^G \hat A_i\left(m_{i,u}-\bar m_u\right),
\qquad
\bar m_u=\frac{1}{G}\sum_{i=1}^G m_{i,u}.
\label{eq:advantage_weighted_filter}
\end{equation}
$S_u$ is the covariance between advantage and the presence of direction $u$.
Directions that appear equally in positive and negative rollouts are filtered out, while ones enriched in higher-advantage rollouts survive.
This advantage-weighted filtering is the source of hidden token-level credit assignment in the cancellation view that favours RL training.

\subsection{Cancellation Acts as Hidden Credit Assignment}
\label{sec:hidden_credit_assignment}
Variance reduction and filtering alone does not imply better learning: a smaller update could simply be a weaker update.
The key question is whether cancellation is objective-driven, i.e., whether the residual left after cancellation secretly retains tokens that contribute to the training task.
We compare GRPO with two complementary probes:
\emph{positive-only} keeps only positive-reward rollouts,
\emph{negative-only} keeps only negative-reward rollouts.
Full experimental details are in the Appendix.

\paragraph{Cancellation decreases learning on template and formatting tokens.}
Under the positive-only update, formatting and template tokens dominate the boosted directions, indicating that positive rollouts alone strongly reinforce surface patterns.
By contrast, the joint GRPO update suppresses many of these shared directions and shifts boosted mass toward more reasoning-bearing tokens.
This is the qualitative signature of the filter in Eq.~\ref{eq:advantage_weighted_filter}: directions common across rollout polarities are removed, while directions more correlated with success remain.

\begin{figure}[t]
\vspace{-20pt}
  \centering
    \includegraphics[width=\linewidth]{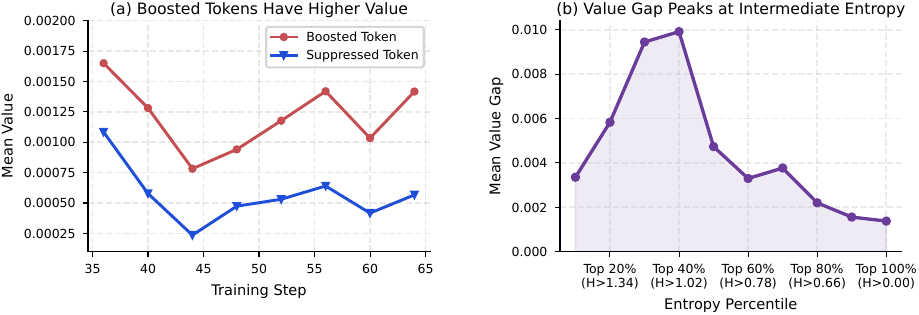}
    \caption{
      Cancellation acts as hidden token-level credit assignment.
      (a) Boosted tokens have higher Monte Carlo estimated value than suppressed tokens.
      (b) The value gap first increases and then decreases with entropy, suggesting that low-entropy tokens dilute the learning signal.
      }
  \label{fig:value_combined_row}
\vspace{-15pt}
\end{figure}

\paragraph{Cancellation favors tokens with higher value.}
We further ask whether the tokens that survive cancellation are more critical.
We thus estimate token value with Monte Carlo rollouts.
We randomly sample equal numbers of boosted and suppressed tokens from positive or negative rollouts, and for each sampled token with state $s_t=(q,o_{<t})$ and action $a_t=o_t$, we estimate the marginal contribution of $a_t$ by rolling out $M$ continuations from both $s_t\oplus a_t$ and $s_t$:
\[
\hat A_M(s_t,a_t)
\;=\;
\frac{1}{M}\sum_{m=1}^{M} r\!\left(\tau^{(m)} \mid s_t \oplus a_t\right)
\;-\;
\frac{1}{M}\sum_{m=1}^{M} r\!\left(\tilde{\tau}^{(m)} \mid s_t\right),
\]
where $r\in\{0,1\}$ is the binary reward.
As shown in the Fig.~\ref{fig:value_combined_row}(a), boosted tokens consistently have higher Monte Carlo estimated value than suppressed tokens, indicating that cancellation does not merely remove surface noise.

\paragraph{The value gap peaks at intermediate entropy.}
To examine where cancellation carries the most informative signal, we sort all tokens by entropy and compute the mean value gap on the top-$k\%$ highest-entropy subset for varying $k$.
As shown in Fig.~\ref{fig:value_combined_row}(b), the gap peaks at about top-40\%.
The trend reflects that high entropy could amplify the value gap.
And low-entropy tokens, anchored by pretraining priors, resist displacement and contribute little net signal, diluting the average gap.
The observation connects to recent high-entropy token update strategies~\citep{wang2025beyond, zheng2025first, wu2025spine}.
At the same time, the rise-then-fall pattern suggests that entropy is not a monotonic proxy for token value; the strongest hidden credit signal appears before low-entropy tokens are included, rather than simply at the maximum-entropy extreme.

\begin{wrapfigure}{r}{0.47\textwidth}
  \vspace{-1.2em}
  \centering
  \includegraphics[width=\linewidth]{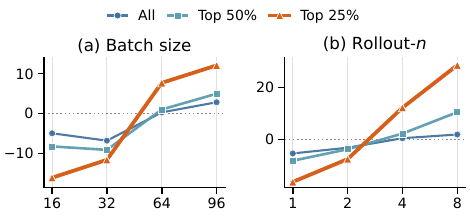}
  \vspace{-0.8em}
  \caption{
    Cancellation strengthens with the sampling budget. The value gap ($10^{-3}$) grows as
    batch size increases from 16 to 96 (a) and rollout-$n$ from 1 to 8 (b),                                                                      
    with the effect most pronounced for higher-value token subsets.
  }
  \label{fig:value_bs}
  \vspace{-1.0em}
\end{wrapfigure}

\paragraph{Larger sampling budget strengthens cancellation.}
Cancellation relies on the co-occurrence of positive and negative rollouts within each update.
When the sampling budget is small, many query groups lack sufficient rollout diversity, and the opposing coupled signals that drive cancellation cannot fully form.
We verify this by independently varying batch size and rollout-$n$, while keeping other hyperparameters fixed.
As shown in Fig.~\ref{fig:value_bs}, the results are consistent across both axes: the hidden value discovery grows with larger sampling budget.
Notably, this amplification is most pronounced for the top-25\% highest-value subset.
This confirms that larger batches yield more complete coupling groups, strengthening cancellation and amplifying the value signal they expose.

\section{Repairing Broken Cancellation in RLVR}
\label{sec:repair}
The previous section suggests that cancellation is not merely an analytical artifact: it reduces shared update directions and exposes a more selective token-level learning signal. We are then motivated to investigate whether common practices in RL frameworks inadvertently break this mechanism and whether repairing it yields consistent performance improvements.

\paragraph{Query-preserved mini-batching}
To improve training throughput, most RL frameworks often split rollout batches \emph{randomly} into smaller mini-batches, typically by token count~\citep{verl,slime}.
This implementation detail could scatter rollouts from the same query across different optimizer steps.
From the cancellation view, this could be harmful, as the positive and negative rollouts of a query are better updated together to reflect their shared directions for cancellation.

(1) The first issue is local imbalance.
Consider a query group with group-normalized advantages $\{\hat A_i\}_{i=1}^G$ and $\sum_i \hat A_i=0$.
If only a subset $\mathcal{B}$ appears in a mini-batch, its local advantage sum
$S_{\mathcal{B}}=\sum_{i\in\mathcal{B}}\hat A_i$
is generally nonzero.
Consequently, the local cross-advantage term becomes
\begin{equation}
    \sum_{i\ne j\in\mathcal{B}}\hat A_i\hat A_j
    =
    S_{\mathcal{B}}^2-\sum_{i\in\mathcal{B}}\hat A_i^2,
\end{equation}
rather than the negative zero-sum form in the grouped update.
Thus, the negative cross terms that reduce variance in Sec.~\ref{sec:structured_cancellation} are weakened or even lost, making the local update noisier.
(2) The second issue is optimization drift.
One might expect later mini-batches to provide the missing counterparts, but PPO/GRPO updates the policy after every mini-batch.
Therefore, later rollouts use shifted score directions and different importance-sampling ratios.
The sequential mini-batch trajectory is not equivalent to the simultaneous grouped update, so cancellation missing in an early mini-batch cannot be exactly restored later.
We thereby implement query-preserved mini-batching, ensuring that rollouts for the same query are grouped into a single mini-batch and updated together.

\begin{wrapfigure}{r}{0.6\textwidth}
  \vspace{-1.0em}
  \centering
  \includegraphics[width=0.6\textwidth]{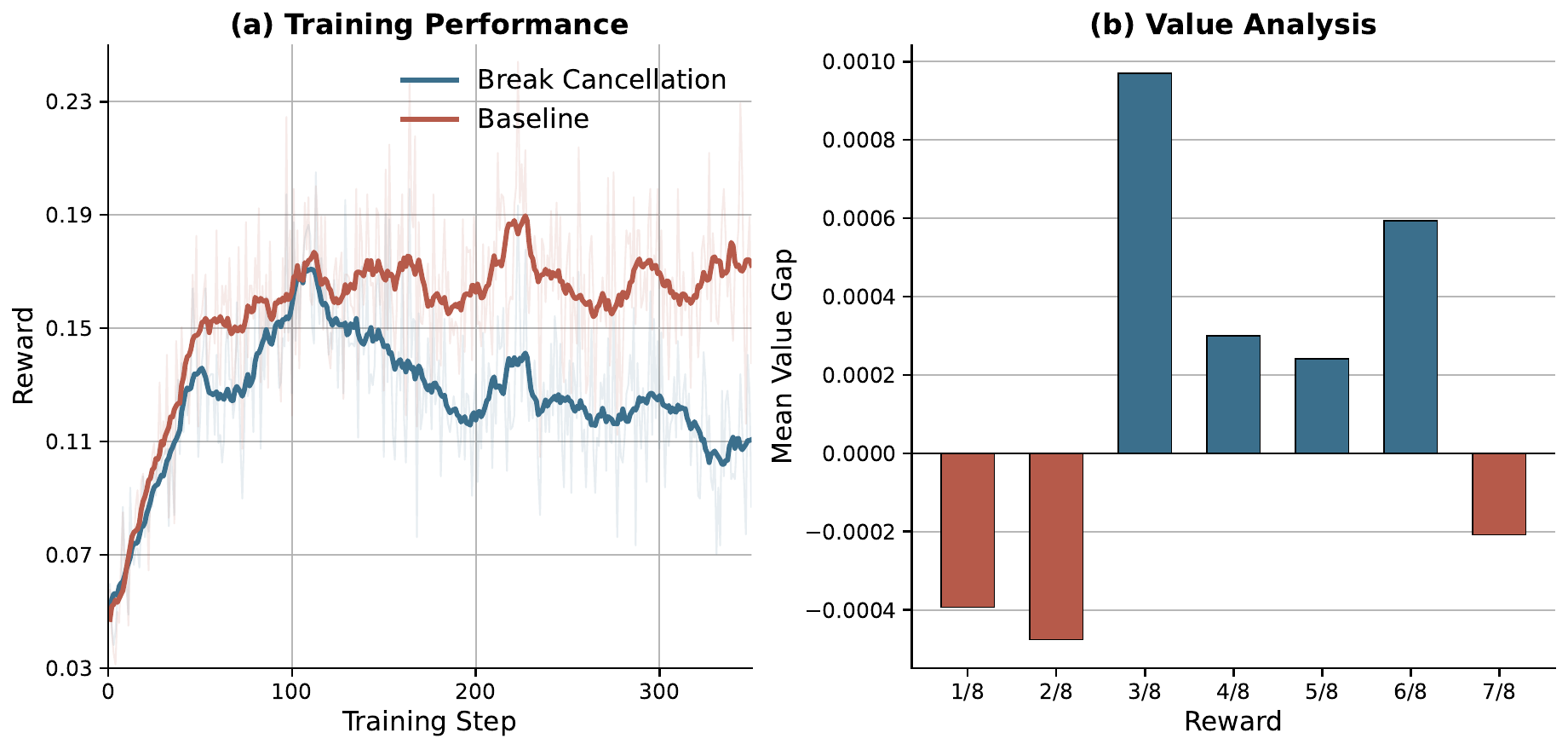}
  \caption{(a) Partitioning mini-batches by reward sign maximally breaks cancellation, causing substantial performance drops. (b) Value gap between boosted and suppressed tokens across reward distributions: Gap diminishes when reward is sparse or saturated}
  \label{fig:imbalance}
  \vspace{-1.0em}
\end{wrapfigure}

\paragraph{Reward-balanced batching}

As analyzed in section~\ref{sec:structured_cancellation}, the cancellation mechanism relies on the co-occurrence of positive and negative advantage within the same update.
When batch reward is sparse or saturated, the batch may lack sufficient cross-sign counterparts, and the coupled signals that drive cancellation cannot fully form. 
We verify this failure mode through two complementary experiments:
(1) We partition each sampling batch by advantage sign, placing positive and negative rollouts into separate mini-batches.
This removes cross-sign gradient coupling, deliberately breaking cancellation and the training performance degrades.
(2) We measure the value gap between boosted and suppressed tokens, varying the reward-sign composition within the same query groups.
As shown in~\ref{fig:imbalance}, under reward-sparse or saturated conditions, the value gap diminishes or even becomes negative.

These observations motivate a simple remedy: preserve reward-sign balance during batch construction.
Therefore, we accumulate rollouts until a reward-balanced subset can be selected for update.
Specifically, we accumulate rollouts and defer updates until both reward signs are sufficiently represented.
We then select an update batch $\mathcal{B}$ satisfying
\[
\min\{N_+(\mathcal{B}), N_-(\mathcal{B})\} \geq \tau |\mathcal{B}|,
\]     
where $N_+(\mathcal{B})$ and $N_-(\mathcal{B})$ denote the numbers of positive and negative rollouts in $\mathcal{B}$, and $\tau$ controls the minimum fraction of either sign.

\paragraph{Experiment results}

\begin{figure}[t]
\vspace{-30pt}
  \centering
  \includegraphics[width=\linewidth]{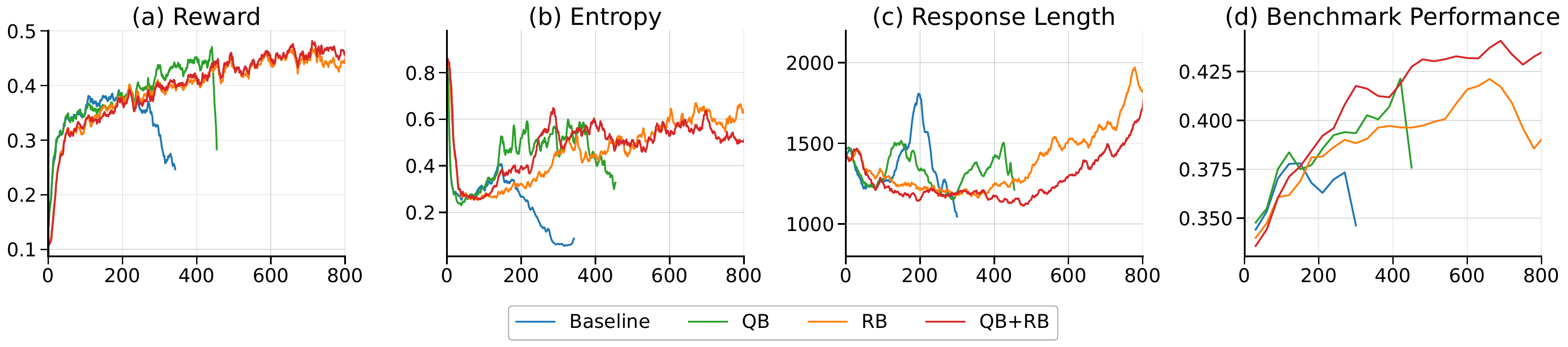}
  \caption{
    Query-preserved mini-batching (QB) and reward-balanced batching (RB) stabilize GRPO training on Qwen2.5-Math-7B. The baseline suffers severe training collapse, while QB and RB individually mitigate instability. Their combination (QB+RB) achieves the most stable training dynamics and the strongest final performance across all evaluation benchmarks.
  }
  \label{fig:evaluation_7b}
  \vspace{-10pt}
\end{figure}

Full experiment details can be found in Appendix~\ref{appendix:experiment}.
Table~\ref{tab:method-combination-results} presents the evaluation results across all model scales and benchmarks.
Our proposed interventions achieve consistent improvements, with particularly notable results on the AMC and Olympiad Bench.
This phenomenon is most pronounced for Qwen2.5-Math-7B, which exhibits the most unstable training when RL is from base models.
As shown in Figure~\ref{fig:evaluation_7b}, the baseline configuration suffers from severe collapse, whereas batching strategies stabilize the training and improve final performance.

\begin{table*}[t]
\small
  \centering
  \caption{
  Evaluation results across math reasoning benchmarks under different method combinations (\%).
  QB: query-preserved mini batching
  RB: reward-balanced batching
  }
  \label{tab:method-combination-results}

  \small
  \setlength{\tabcolsep}{4.2pt}
  \renewcommand{\arraystretch}{1.10}

  \begin{tabular*}{\textwidth}{
    @{\extracolsep{\fill}}
    p{1.8cm}
    cc
    *{7}{S[table-format=2.1]}
    @{}
  }
    \toprule
    \textbf{Model} & \multicolumn{2}{c}{\textbf{Methods}} & \multicolumn{7}{c}{\textbf{Benchmarks}} \\
    \cmidrule(lr){2-3}
    \cmidrule(lr){4-10}
    & \textbf{QB} & \textbf{RB}
    & \textbf{AIME24} & \textbf{AIME25} & \textbf{AMC} & \textbf{Olympiad} & \textbf{Minerva} & \textbf{Math} & \textbf{Avg} \\
    \midrule
    \multirow{4}{=}{\makecell[l]{Qwen2.5-\\Math-7B}}
    &        &        & 21.3 & 11.3 & 54.9 & 37.8 & 29.4 & 72.6 & 37.8 \\
    & \cmark &        & 20.7 & 10.8 & 61.2 & 42.6 & \textbf{36.7} & 81.6 & 42.3 \\
    &        & \cmark & 25.4 & 13.3 & 60.1 & 39.0 & 32.4 & 73.0 & 42.4 \\
    & \cmark & \cmark & \textbf{27.8} & \textbf{15.9} & \textbf{63.9} & \textbf{44.5} & 36.0 & \textbf{82.2} & {\bfseries 45.0} \\
    \midrule

    \multirow{4}{=}{\makecell[l]{Qwen3-1.7B\\-Base}}
    &        &        & 12.4 &  9.4 & 37.0 & 34.5 & 28.3 & 69.8 & 31.9 \\
    & \cmark &        & 11.6 &  7.9 & 40.2 & 37.6 & 33.8 & 71.0 & 33.7 \\
    &        & \cmark & 16.2 & \textbf{11.8} & 42.6 & 37.3 & 30.9 & \textbf{74.2} & 35.5 \\
    & \cmark & \cmark & \textbf{16.5} & 9.0 & \textbf{43.5} & \textbf{38.2} & \textbf{35.7} & 74.0 & {\bfseries 36.1} \\
    \midrule

    \multirow{4}{=}{\makecell[l]{Qwen3-4B\\-Base}}
    &        &        & 28.3 & \textbf{24.2} & 58.0 & 53.2 & 44.5 & 86.2 & 49.1 \\
    & \cmark &        & 29.2 & 22.8 & \textbf{64.4} & 53.6 & 44.9 & 86.4 & 50.2 \\
    &        & \cmark & 27.9 & 22.4 & 61.5 & 54.8 & \textbf{46.0} & 88.0 & 50.1 \\
    & \cmark & \cmark & \textbf{30.7} & 23.0 & 64.0 & \textbf{55.4} & 43.4 & \textbf{88.4} & \textbf{50.8} \\
    \bottomrule
  \end{tabular*}
\end{table*}

\vspace{-8pt}
\section{Related Work}
A common sequence-level intuition for LLM RL training is that it reinforces successful rollouts and suppresses failed ones, shifting the model distribution toward higher-reward responses~\citep{mroueh2025reinforcement, wu2025takes}.
However, recent studies suggest that this explanation could be incomplete.
The \emph{Lazy Likelihood} phenomenon shows that positively supervised responses are not necessarily upweighted during training~\citep{razin2024unintentional,pal2024smaug,ren2024learning,deng2025effect,deng2025token}.
Related positive-suppressed phenomena have also been reported across diverse settings of LLM post-training, including reasoning with format-reward~\citep{simoni2025gtpo}, search-integrated RL~\citep{deng2025grpo}, and multi-turn tool-integrated reasoning~\citep{xue2025simpletir}. 
Existing studies primarily focused on domain-specific remedies for particular instability modes the \emph{Lazy Likelihood} may cause, or on rollout-level analyses of likelihood dynamics, while leaving the in-depth analysis of how individual tokens are updated in critic-free RL remains underexplored.

\vspace{-8pt}
\section{Conclusion} 
In this work, we revisit critic-free RL for LLMs from the perspective of token-level probability shifts.
We identified \emph{token flipping}, showing that positive and negative rollouts exhibit remarkably similar proportions of boosted and suppressed tokens, despite receiving opposite signs of advantage.
To explain this mismatch, we derived a first-order decomposition of GRPO updates, showing that a token's log-probability change is shaped by a coupling kernel that concentrates on same-token and low-confidence token pairs.
Building on this view, we propose the \emph{cancellation hypothesis}: opposing coupled signals cancel out for tokens shared across rollouts, while tokens more specific to successful ones receive stronger reinforcement, thereby inducing hidden token-level credit assignment from outcome-level supervision.
Guided by this view, we further introduce two batching interventions for RL training and validate their performance.
We hope this work offers a new perspective on the learning dynamics of critic-free RL for LLMs and will pave the way for more principled approaches to batch design and credit assignment in LLM post-training.

\nocite{*}
\bibliographystyle{plainnat}
\bibliography{main}

\begin{ack}
IT acknowledges support from Dutch National Science Foundation (NWO Vici grant VI.C.212.053).
\end{ack}

\appendix

\input{appendix}


\end{document}

%% file: appendix.tex
\section{Limitations}
Our coupling analysis relies on an output-layer proxy that approximates the full-model gradient with its last-layer counterpart, so we validate the reasonableness of this proxy and provide corresponding empirical justifications: Appendix~\ref{appendix:proxy} shows that the proxy and full-parameter updates exhibit stable token-flip sign consistency throughout training, while Table~\ref{tab:directional_masking} shows that the same-token, low-confidence coupling pattern identified by the proxy produces comparable masking effects under full-parameter updates.
Our two interventions are evaluated on mathematical reasoning with binary rewards. Extending them and validating them in more diverse settings and broader domains remains future work.
Reward-balanced batching may reduce sample utilization because some rollouts are discarded to meet the target reward composition. However, our design naturally fits modern asynchronous RL frameworks: it can be implemented simply as an additional update condition on the rollout buffer.

\section{Experiment Details}
\label{appendix:experiment}

\paragraph{Experiment Settings} We employ math reasoning problems as a playground to test the proposed method, for math problems usually have reliable and cheap verifiers. Our training set is the publicly released DAPO-17K-dataset. Each training batch consists of 128 prompts with 8 responses sampled per prompt, yielding a total of 1024 samples. In our configuration, updates are performed with a mini-batch of 256 samples, resulting in 4 gradient updates per sampling iteration.
We evaluate our analysis on three backbone models spanning both math-specialized and     
general-purpose LLMs: \textsc{Qwen2.5-Math-7B}, \textsc{Qwen3-1.7B-Base}, and            
\textsc{Qwen3-4B-Base}. This model suite allows us to examine whether the observed       
token-level dynamics remain consistent across different model families and scales. We use
 a maximum response length of 8192 for all models. We use the DAPO dataset throughout our
 experiments and adopt \textsc{Math-Verify} as the verifier.

\paragraph{Implementation} 
We implement all RL experiments in the \textsc{verl} framework and use GRPO as our critic-free RL algorithm.
Following DAPO~\citep{yu2025dapo}, we adopt the token-level policy gradient loss and enable \texttt{clip-higher}.
However, we do not use Dynamic Sampling due to limited computational budget.  
In all experiments, we set $\texttt{clip\_low}=0.2$ and $\texttt{clip\_high}=0.28$, remove the KL regularization term by setting $
\beta=0$, and set the entropy loss coefficient to $0$. 
We use Math-Verify as our reward function and include no format or length reward, our training algorithm minimizes: 
\begin{equation}
  \mathbb{E}_{q \sim \mathcal{D}, \{o_i\}_{i=1}^G \sim \pi_{\theta_{\text{old}}}} \left[ \frac{1}{\sum_{i=1}^G |o_i|} \sum_{i=1}^G \sum_{t=1}^{|o_i|} \min \left( \rho_{i,t}(\theta) A_i, \text{clip}(\rho_{i,t}(\theta), 1-\epsilon, 1+\epsilon) A_i \right) \right]
\end{equation}
where $\epsilon$ is the clipping parameter and $\rho_{i,t}(\theta) = \frac{\pi_{\theta}(o_{i,t}|q, o_{i,<t})}{\pi_{\theta_{\text{old}}}(o_{i,t}|q, o_{i,<t})}$ is the importance ratio. Table~\ref{tab:rl-practice} summarizes the default training configurations.
\begin{table}[h]
  \caption{Training hyperparameters across different models.}
  \label{tab:rl-practice}
  \centering
  \small
  \setlength{\tabcolsep}{6pt}
  \renewcommand{\arraystretch}{1.08}
  \resizebox{\linewidth}{!}{%
  \begin{tabular}{l|ccc}
  \toprule
  \textbf{Hyperparameters} & \textbf{Qwen2.5-Math-7B} & \textbf{Qwen3-1.7B-Base} & \textbf{Qwen3-4B-Base} \\
  \midrule
  \texttt{max\_prompt\_length} & 1024 & 1024 & 1024 \\
  \texttt{max\_response\_length} & 8192 & 8192 & 8192 \\
  \texttt{train\_batch\_size} & 128 & 128 & 128 \\
  \texttt{mini\_batch\_size} & 32 & 32 & 32 \\
  \texttt{optim.lr} & $1\mathrm{e}{-6}$ & $1\mathrm{e}{-6}$ & $1\mathrm{e}{-6}$ \\
  \texttt{rollout.temperature} & 1.0 & 1.0 & 1.0 \\
  \texttt{rollout.n} & 8 & 8 & 8 \\
  \bottomrule
  \end{tabular}%
  }
\end{table}

\paragraph{Implementation Details for proposed batching tricks} We implement all experiments within the \textsc{verl} framework, a widely adopted
open-source RL training framework.  By default, \textsc{verl} partitions mini-batches by token count to balance computational
load across data-parallel nodes. For \textit{query-preserved mini-batching}, we ensure all rollouts from the same query
are assigned to the same mini-batch.   For \textit{reward-balanced batching}, we accumulate rollouts in a buffer and trigger
  gradient updates only when the collected batch meets the target reward composition.      
  We set $\tau=0.5$ for \textsc{Qwen2.5-Math-7B} and \textsc{Qwen3-4B-Base}, and
  $\tau=0.25$ for \textsc{Qwen3-1.7B-Base}.

\paragraph{Evaluation} For evaluation, we mainly focus on six widely used math reasoning benchmarks, including AIME2024, AIME2025, AMC, Minerva, OlympiadBench and Math-500. For AIME2024, AIME2025, and AMC, we report avg@32 because the test set is relatively small; for the other three benchmarks, we report pass@1.
To determine the evaluation hyperparameters, we test four common sampling configurations on both the base and aligned models, as shown in Table~\ref{tab:sampling-results}. We adopt temperature $= 0.6$ and top-$p = 0.95$ for all evaluations, as this setting yields the highest average performance on the aligned model while maintaining competitive results on base model.

\begin{table}[h]
  \centering
  \caption{Evaluation results under different temperature and top-$p$ settings.}
  \label{tab:sampling-results}
  \small
  \begin{tabular}{lccrrrrrrr}
  \toprule
  Model & Temp. & Top-$p$ & AIME24 & AIME25 & AMC & Math & Olymp. & Minerva & Avg. \\
  \midrule
  qwen3-4b-base & 1.0 & 0.7  & \textbf{10.6} &  6.7 & 36.4 & 62.8 & 33.4 & 24.2 & 29.0 \\
  qwen3-4b-base & 0.6 & 0.7  & 10.0 &  \textbf{8.5} & \textbf{41.7} & \textbf{73.6} & \textbf{36.7} & 26.1 & \textbf{32.8} \\
  qwen3-4b-base & 1.0 & 0.95 &  6.6 &  3.6 & 27.1 & 49.0 & 23.8 & 17.6 & 21.3 \\
  qwen3-4b-base & 0.6 & 0.95 & 10.2 &  6.6 & 39.9 & 64.8 & 34.5 & \textbf{27.9} & 30.7 \\
  \midrule
  hf-4b-step100 & 1.0 & 0.7  & 16.0 & 14.1 & 49.5 & \textbf{82.6} & 46.2 & \textbf{37.5} & 41.0 \\
  hf-4b-step100 & 0.6 & 0.7  & 14.2 & 12.3 & 49.9 & 81.0 & \textbf{46.9} & 37.1 & 40.2 \\
  hf-4b-step100 & 1.0 & 0.95 & \textbf{16.8} & 13.9 & 49.4 & 79.8 & 44.4 & \textbf{37.5} & 40.3 \\
  hf-4b-step100 & 0.6 & 0.95 & 15.8 & \textbf{14.8} & \textbf{50.6} & 82.4 & \textbf{46.9} & 36.7 & \textbf{41.2} \\
  \bottomrule
  \end{tabular}
\end{table}

\section{Token Flipping}

\paragraph{Measuring token-level log-probability shift.}
We measure how a single GRPO update changes token probabilities on a fixed rollout batch.
In typical RL training pipelines, responses are sampled by a fast inference engine as VLLM, which also return rollout-time log-probabilities. 
However, these values can differ slightly from those produced by the FSDP training forward pass, even for the same model state, due to numerical differences between the inference and training stacks.
To ensure an exact comparison, we compute both the pre-update and post-update log-probabilities via dedicated FSDP forward passes on the training model over the same rollout batch. 
Concretely, before the GRPO update we obtain $\log\pi_\theta(o_{i,t}\mid q,o_{i,<t})$, and after the update we obtain $\log\pi_{\theta'}(o_{i,t}\mid q,o_{i,<t})$ through a second FSDP forward pass. The per-token shift and its classification are:
\begin{gather}
  \Delta \log p_{i,t} = \log\pi_{\theta'}(o_{i,t}\mid q,o_{i,<t}) - \log\pi_{\theta}(o_{i,t}\mid q,o_{i,<t}), \\
  \textit{boosted}:\; \Delta \log p_{i,t} > \epsilon, \qquad
  \textit{suppressed}:\; \Delta \log p_{i,t} < -\epsilon, \qquad
  \textit{stable}:\; |\Delta \log p_{i,t}| \le \epsilon, \notag
\end{gather}
with $\epsilon=10^{-6}$. We report the ratio of boosted and suppressed tokens separately for positive-advantage and negative-advantage rollouts. The full procedure is summarized in Algorithm~\ref{alg:delta_logprob}.

\begin{algorithm}[h]
\caption{Computing token-level $\Delta \log p$ after one GRPO step.}
\label{alg:delta_logprob}
\begin{algorithmic}[1]
\Require Batch of prompts $\{q\}$, current policy $\pi_\theta$, number of rollouts $G$
\Statex \hspace{-1.2em}\textcolor{gray}{\hrulefill~\textit{Rollout \& Scoring}~\hrulefill}
\State For each prompt $q$, sample $G$ responses $\{o_i\}_{i=1}^G \sim \pi_\theta(\cdot \mid q)$ \hfill {\footnotesize\textcolor{gray}{via vLLM}}
\State Compute $\ell^{\text{old}}_{i,t} \gets \log \pi_\theta(o_{i,t} \mid q, o_{i,<t})$ for all response tokens \hfill {\footnotesize\textcolor{gray}{via FSDP forward}}
\State Verify each response; compute group-normalized advantage $A_i$
\Statex \hspace{-1.2em}\textcolor{gray}{\hrulefill~\textit{Policy Update}~\hrulefill}
\State $\theta' \gets \textsc{GRPO-Update}(\theta, \{o_i, A_i\})$
\Statex \hspace{-1.2em}\textcolor{gray}{\hrulefill~\textit{Measure Token Displacement}~\hrulefill}
\State Compute $\ell^{\text{new}}_{i,t} \gets \log \pi_{\theta'}(o_{i,t} \mid q, o_{i,<t})$ for all response tokens \hfill {\footnotesize\textcolor{gray}{via FSDP forward}}
\State $\Delta \log p_{i,t} \gets \ell^{\text{new}}_{i,t} - \ell^{\text{old}}_{i,t}$
\end{algorithmic}
\end{algorithm}

\begin{figure}[h]
  \centering
  \includegraphics[width=1\linewidth]{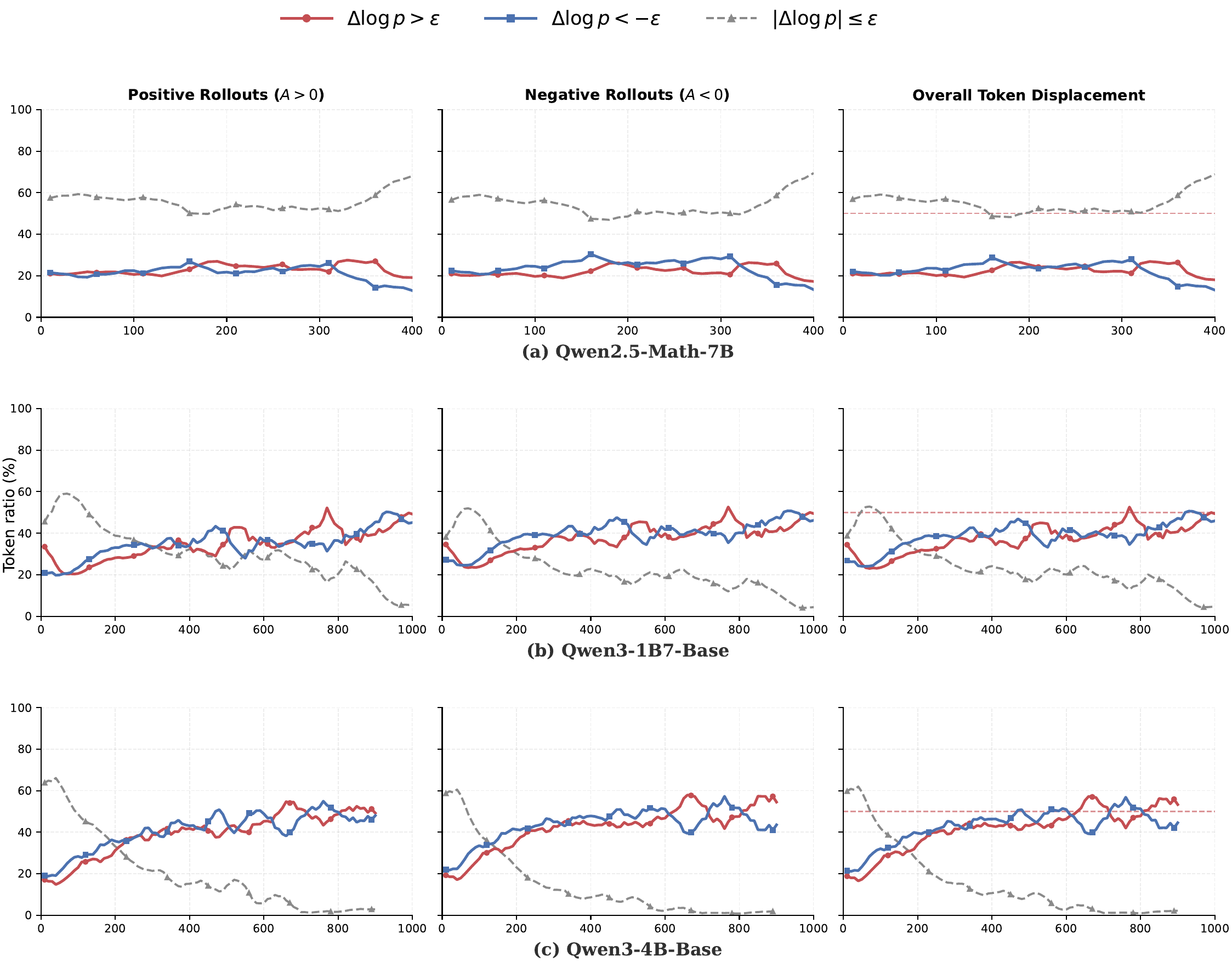}
  \caption{\textbf{Reversed update persists across backbone models.} Each row corresponds to a backbone model, and
  the three columns report the proportions of boosted ($\Delta \log p > \epsilon$), suppressed ($\Delta \log p <
  -\epsilon$), and stable ($|\Delta \log p| \le \epsilon$) tokens in positive rollouts ($A>0$), negative rollouts
  ($A<0$), and the full rollout batch. Despite opposite advantage signs, positive and negative
  rollouts exhibit remarkably similar token-displacement profiles across models.}
  \label{fig:reversed-update-overview}
\end{figure}

\paragraph{Additional empirical evidence for token flipping.}
Figure~\ref{fig:reversed-update-overview} shows that \emph{Token Flipping} is a stable phenomenon across different models. 
We report the proportions of boosted, suppressed, and stable tokens, defined by $\Delta \log p$, for positive rollouts ($A>0$), negative rollouts ($A<0$), and the full rollout batch. 
Across different models, Positive and negative rollouts display highly similar token-displacement profiles throughout generation, even though their advantages have opposite signs.

\section{Coupling Analysis}
\subsection{Empirical Justification for the Output-Layer Proxy}
\label{appendix:proxy}
The coupling kernel in Eq.~\eqref{eq:coupling_main} is defined in the full parameter space, making it expensive to compute and difficult to interpret. This motivates a simpler proxy based on the output layer. We provide empirical evidence that this proxy remains faithful to the full-parameter update in practice.

\begin{figure}[htbp]
    \centering
    \begin{subfigure}{0.7\linewidth}
        \centering
        \includegraphics[width=\linewidth]{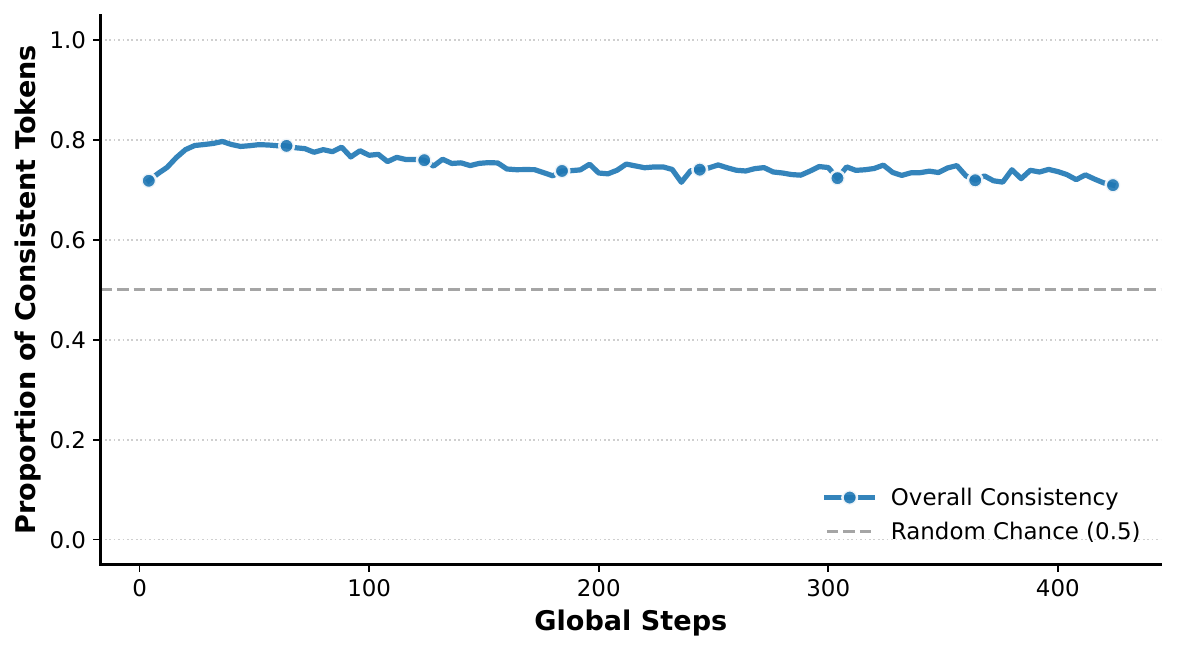}
        \caption{Sign Consistency}
        \label{fig:consistency_sign}
    \end{subfigure}
    \caption{\textbf{Agreement between full-parameter and unembedding-layer updates.} The proportion of tokens where the token displacement ($\Delta \log p$) shares the same sign under both update paradigms. The consistency remains highly stable across training steps, plateauing near $0.7$, providing empirical support for the output-layer proxy adopted in Section~3.}
    \label{fig:layer_proxy_justification}
\end{figure}

To validate the proxy, we compare the per-token displacement $\Delta \log p$ produced by a standard full-parameter GRPO step with the displacement produced by updating only the unembedding matrix $W$, using the same rollout batch.
For each token we record whether its $\Delta \log p$ under the two update paradigms shares the same sign.
As shown in Figure~\ref{fig:layer_proxy_justification}, the sign-consistent proportion remains stable throughout training, plateauing near $0.7$.

Beyond sign consistency, the proxy also faithfully recovers the \emph{structural} predictions of the coupling analysis. Specifically, the factorisation in Eq.~\ref{eq:uncertainty_factorization} predicts that non-negligible coupling concentrates on same-token, low-confidence pairs---a conclusion derived entirely under the output-layer approximation. Table~\ref{tab:directional_masking} provides a direct causal test of this prediction: masking tokens that satisfy both conditions yields comparable Boost Rate and Mean Boost under the LM-head-only and full-parameter updates (${\sim}$74--75\% and ${\sim}$0.030, respectively), whereas relaxing either condition substantially weakens the effect. The close agreement between the two update paradigms confirms that the sparse coupling structure identified by the proxy is not an artifact of the approximation but reflects a property of the full-parameter training dynamics.

\subsection{Empirical Verification Details for Token coupling and Additional Results}

For the masked-update experiments reported in Table~\ref{tab:directional_masking} and Figure~\ref{fig:causal_study}, we deliberately use SGD with a learning rate of $1\mathrm{e}{-1}$ and no momentum rather than Adam.
The reason is that Adam's first parameter update reduces to an approximate sign-descent step (see Appendix~\ref{app:adam-sign} for a formal derivation), which normalizes away the relative magnitude differences across tokens and would obscure the per-token causal signal our analysis aims to detect.
Following~\citet{mukherjee2026we}, we adopt SGD for RL updates with a learning rate of $1\mathrm{e}{-1}$, which preserves the true gradient magnitudes and therefore faithfully reflects the contribution of each masked token set to the candidate update.

As shown in Figure~\ref{fig:causal_study}, the masked-update results provide three complementary pieces of evidence.
(1)~Masking coupled tokens (same-token, low-confidence) produces a clearly right-shifted Masking Effect distribution relative to random tokens (panel~a), indicating that these tokens exert a consistently positive influence on the candidate's log-probability under the unmasked update.
(2)~Boost Rate increases monotonically with coupling strength as measured by the proxy cosine similarity, and the trend is closely matched between the LM-head-only and full-parameter updates (panel~b), further validating the output-layer proxy.
(3)~Progressively masking more top-coupled tokens amplifies both Mean Boost and Boost Rate (panel~c).
Together, these results corroborate the prediction of Eq.~\ref{eq:uncertainty_factorization}: non-negligible coupling is sparse and concentrates on identical, low-confidence token pairs.

\subsection{Why SGD Instead of Adam for Causal Analysis}
\label{app:adam-sign}

In our causal experiments we deliberately use SGD rather than Adam.
The reason is that Adam's first update is governed almost entirely by the \emph{sign} of the gradient, not its magnitude, which would obscure the token-level causal signal we aim to measure.

Concretely, Adam maintains a first-moment estimate $m_t = \beta_1 m_{t-1} + (1-\beta_1) g_t$ and a second-moment estimate $v_t = \beta_2 v_{t-1} + (1-\beta_2) g_t^2$.
At the very first step ($t=1$), with $m_0 = v_0 = 0$, these reduce to
\begin{equation}
  m_1 = (1-\beta_1)\,g_1, \qquad v_1 = (1-\beta_2)\,g_1^2.
\end{equation}
After bias correction ($\hat{m}_1 = m_1/(1-\beta_1)$, $\hat{v}_1 = v_1/(1-\beta_2)$), the parameter update becomes
\begin{equation}
  \theta_1 = \theta_0 - \alpha \cdot \frac{\hat{m}_1}{\sqrt{\hat{v}_1} + \epsilon}
            = \theta_0 - \alpha \cdot \frac{g_1}{\sqrt{g_1^2} + \epsilon}
            = \theta_0 - \alpha \cdot \frac{g_1}{|g_1| + \epsilon}.
\end{equation}
When $|g_1| \gg \epsilon$ (which holds for the vast majority of parameters), this simplifies to
\begin{equation}
  \theta_1 \approx \theta_0 - \alpha \cdot \operatorname{sign}(g_1).
\end{equation}

This ``sign-descent'' behavior means that the relative contribution of individual tokens to the update is washed out at the first step.
SGD preserves the true gradient magnitudes and therefore faithfully reflects the per-token causal effect, making it the appropriate choice for our analysis.

\section{Cancellation Hypothesis}

\subsection{Decomposing Updates by Reward Sign: Positive-Only and GRPO}

To isolate how positive rollouts contribute to token-level displacement, we construct a controlled
comparison in which the rollout batch is fixed and only the polarity composition of the update is changed.
Given a rollout batch $\mathcal{B}$ with binary rewards $r_i \in \{0,1\}$ and group-normalized advantages $A_i$,
we instantiate two update variants by reweighting the rollout-level advantage:
\begin{itemize}
    \item \textbf{Positive-only:} $A_i^{+} = A_i \cdot \mathbb{I}[r_i = 1]$
    \item \textbf{GRPO (Joint):} use the original $A_i$.
\end{itemize}
Both variants start from the same saved checkpoint $\theta_0$ and operate on the same rollout batch $\mathcal{B}$.
This design removes variation from sampling and initialization, so any difference in the resulting token
displacement can be attributed to the polarity composition of the update itself.
Importantly, after each update, we measure token displacement on the \emph{full original batch} $\mathcal{B}$
rather than on the masked subset used to form the loss.
This shared evaluation protocol makes the displacement maps from Positive-only and GRPO directly comparable.
The full procedure is summarized in Algorithm~\ref{alg:psr_nsr_grpo}.

\begin{algorithm}[h]
\caption{Controlled measurement of token displacement under single-polarity and joint updates.}
\label{alg:psr_nsr_grpo}
\begin{algorithmic}[1]
\Require Dataset $\mathcal{D}$, current policy $\pi_\theta$, number of rollouts $G$
\Statex \hspace{-1.2em}\textcolor{gray}{\hrulefill~\textit{Rollout \& Scoring}~\hrulefill}
\State Sample a batch of prompts $\{q\}$ from $\mathcal{D}$
\State For each prompt $q$, sample $G$ responses $\{o_i\}_{i=1}^G \sim \pi_\theta(\cdot \mid q)$
\State Verify each rollout to obtain reward $r_i \in \{0,1\}$; compute group-normalized advantage $A_i$
\State Collect rollout batch $\mathcal{B} = \{(q, o_i, r_i, A_i)\}$
\Statex \hspace{-1.2em}\textcolor{gray}{\hrulefill~\textit{Shared Preparation}~\hrulefill}
\State Compute $\ell^{\text{old}}_{i,t} \gets \log \pi_\theta(o_{i,t} \mid q, o_{i,<t})$ for all tokens in $\mathcal{B}$
\State Save checkpoint $\theta_0 \gets \theta$
\Statex \hspace{-1.2em}\textcolor{gray}{\hrulefill~\textit{Positive-Only Update}~\hrulefill}
\State \hspace{1.0em} Define $A_i^{+} = A_i \cdot \mathbb{I}[r_i = 1]$ for all rollouts
\State \hspace{1.0em} $\theta^{+} \gets \textsc{GRPO-Update}(\theta_0,\; \mathcal{B},\; \{A_i^{+}\})$
\State \hspace{1.0em} $\Delta \log p^{\textsc{pos}}_{i,t} \gets \log \pi_{\theta^{+}}(o_{i,t} \mid q, o_{i,<t}) - \ell^{\text{old}}_{i,t}$ \hfill {\footnotesize\textcolor{gray}{measured on full $\mathcal{B}$}}
\Statex \hspace{-1.2em}\textcolor{gray}{\hrulefill~\textit{GRPO: Joint Update}~\hrulefill}
\State \hspace{1.0em} Load checkpoint $\theta \gets \theta_0$
\State \hspace{1.0em} $\theta' \gets \textsc{GRPO-Update}(\theta_0,\; \mathcal{B},\; \{A_i\})$
\State \hspace{1.0em} $\Delta \log p^{\textsc{grpo}}_{i,t} \gets \log \pi_{\theta'}(o_{i,t} \mid q, o_{i,<t}) - \ell^{\text{old}}_{i,t}$ \hfill {\footnotesize\textcolor{gray}{measured on full $\mathcal{B}$}}
\end{algorithmic}
\end{algorithm}

\subsection{Experiment Details for Value Estimation}
In standard RL, the value function is defined for a state rather than for an action:
\[
V^\pi(s_t)
=
\mathbb{E}\big[r \mid s_t,\pi\big].
\]
A token $o_t$ is an action taken at the language-model state $s_t=(q,o_{<t})$,
so it does not have a context-free value by itself. The standard action-related
quantity is instead the state-action Q fuction:
\[
Q^\pi(s_t,o_t)
=
\mathbb{E}\big[r \mid s_t \oplus o_t, \pi\big],
\]
which measures the expected reward after forcing $o_t$ and then following policy $\pi$.

Our analysis focuses on a different quantity: at the same state $s_t$, how much
better is taking token $o_t$ than not taking it and instead sampling an
alternative token from the policy? We define this \emph{counterfactual token value} as
\[
\Delta^\pi(o_t \mid s_t)
=
Q^\pi(s_t,o_t)-Q^\pi_{\neg o_t}(s_t),
\]
where
\[
Q^\pi_{\neg o_t}(s_t)
=
\mathbb{E}_{a\sim \pi(\cdot\mid s_t),\, a\neq o_t}
Q^\pi(s_t,a)
\]
is the expected value of the counterfactual alternatives.

Let $p_t=\pi(o_t\mid s_t)$. The state value mixes the forced-token branch and
the alternatives:
\[
V^\pi(s_t)
=
p_t Q^\pi(s_t,o_t)
+
(1-p_t)Q^\pi_{\neg o_t}(s_t),
\]
Therefore,
\[
Q^\pi(s_t,o_t)-V^\pi(s_t)
=
(1-p_t)
\left(
Q^\pi(s_t,o_t)-Q^\pi_{\neg o_t}(s_t)
\right),
\]
and hence
\[
\Delta^\pi(o_t \mid s_t)
=
\frac{Q^\pi(s_t,o_t)-V^\pi(s_t)}{1-p_t}.
\]

\paragraph{Monte Carlo estimation.}
For each candidate token $o_t$ at state $s_t=(q,o_{<t})$, we estimate the two
terms above with Monte Carlo rollouts. We sample $M{=}256$ complete
continuations from $s_t \oplus o_t$ and denote the mean reward by
$\mathrm{avg}[0]$. We also sample $M{=}256$ complete continuations directly
from $s_t$ and denote the mean reward by $\mathrm{avg}[-1]$. Since rollouts from
$s_t$ already choose $o_t$ with probability $p(o_t)$, the naive difference
$\mathrm{avg}[0]-\mathrm{avg}[-1]$ is attenuated by this self-inclusion. We
therefore use the estimator
\begin{equation}
  \hat{\Delta}(o_t \mid s_t)
  =
  \frac{\mathrm{avg}[0] - \mathrm{avg}[-1]}{1 - p(o_t)}.
\end{equation}
This estimator measures the reward gain of taking token $o_t$ over sampling a
counterfactual alternative from the policy. We refer to this quantity as the
counterfactual token value, or token value for short.

\begin{figure}[h]
  \centering
  \includegraphics[width=1\linewidth]{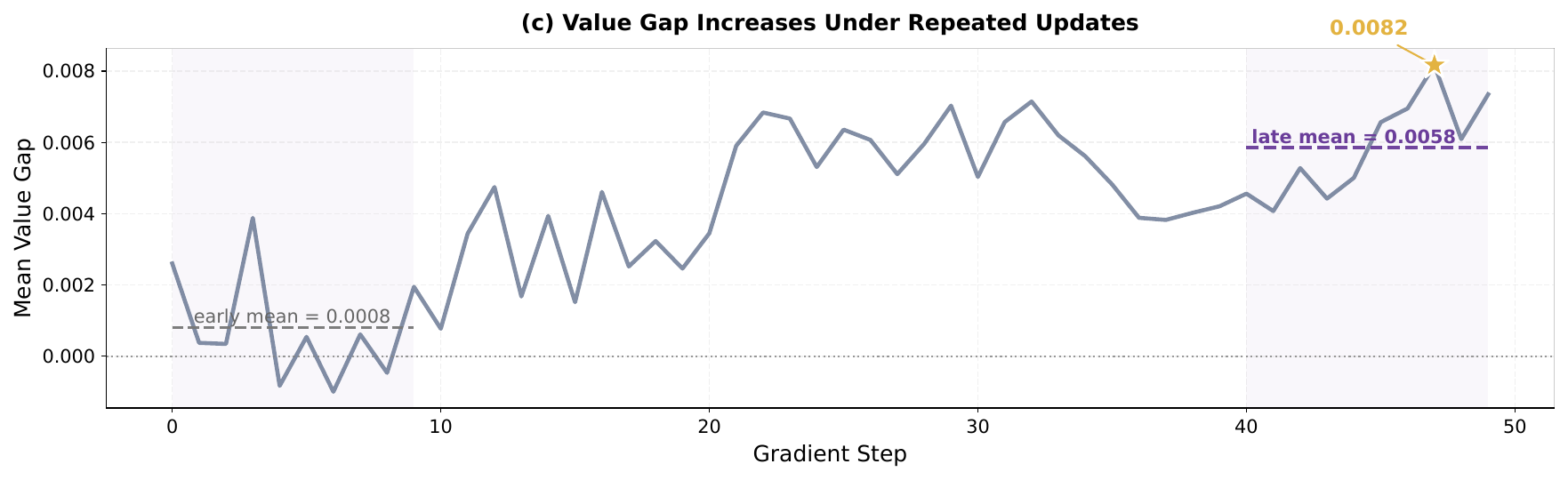}
  \caption{Repeated updates on a fixed batch progressively widen the value gap between 
  boosted and suppressed tokens.}
  \label{fig:repeated-update-value-gap}
\end{figure}

\paragraph{Repeated Updates widen value gap}
We apply repeated GRPO updates on the same fixed batch and track the mean value gap between boosted and suppressed tokens across update steps. As shown in Figure~\ref{fig:repeated-update-value-gap}, the value gap grows monotonically 
with the number of updates, indicating that each gradient step further separates         
high-value boosted tokens from low-value suppressed ones.                                
This accumulation suggests that the hidden credit assignment embedded in cancellation is 
not a one-step coincidence but a stable structural property of critic-free RL: successive
updates consistently surface and amplify the implicit value signal latent in the rollout 
batch. 

